\documentclass{article}

\usepackage{arxiv}

\usepackage[utf8]{inputenc} % allow utf-8 input
\usepackage[T1]{fontenc}    % use 8-bit T1 fonts
\usepackage{hyperref}       % hyperlinks
\usepackage{url}            % simple URL typesetting
\usepackage{booktabs}       % professional-quality tables
\usepackage{amsfonts}       % blackboard math symbols
\usepackage{nicefrac}       % compact symbols for 1/2, etc.
\usepackage{microtype}      % microtypography
\usepackage{lipsum}

\usepackage{amsmath,amssymb} 

\usepackage{paralist}
\usepackage[inline,shortlabels]{enumitem}
\usepackage{multirow}
\usepackage{adjustbox}

\title{WSRNet: Joint Spotting and Recognition of Handwritten Words}

\author{
  George Retsinas\\
  School of Electrical and Computer Engineering, National Technical University of Athens, Greece\\
  \texttt{gretsinas@central.ntua.gr} \\
   \And
 Giorgos Sfikas\\
  Dpt. of Computer Science and Engineering, University of Ioannina,  Greece\\
  \texttt{sfikas@cs.uoi.gr} \\
  \And
 Petros Maragos\\
  School of Electrical and Computer Engineering, National Technical University of Athens, Greece\\
  \texttt{maragos@cs.ntua.gr} \\
}

\begin{document}
\maketitle

\begin{abstract}
  In this work, we present a unified model that can handle both Keyword Spotting and Word Recognition with the same network architecture.
  The proposed network is comprised of a non-recurrent CTC branch and a Seq2Seq branch that is further augmented with an Autoencoding module.
  The related joint loss leads to a boost in recognition performance, while the Seq2Seq branch is used to create efficient word representations.
  We show how to further process these representations with binarization and a retraining scheme to provide compact and highly efficient descriptors, suitable for keyword spotting.
  Numerical results validate the usefulness of the proposed architecture, as our method outperforms the previous state-of-the-art in keyword spotting, and provides results in the ballpark of the leading methods for word recognition.
%Papers must not be larger than 30MB. 
%Papers with more than 14 pages (excluding references), or that use a formatting different from the provided template, will be rejected without review.
\end{abstract}

\section{Introduction}

Handwritten Text Recognition (HTR) and Keyword Spotting (KWS) are two tasks of central importance in the literature of document image processing.
HTR deals with automatic transcription of segmented lines of text \cite{fischer2012handwriting} or isolated words \cite{poznanski2016cnn,sfikas2017phoc}, while in keyword spotting the goal is to detect instances of specific keyword in a given digitized document.
Keyword spotting may be desirable over full text recognition, especially in cases where recognition of the whole text is unnecessary or would likely be suboptimal \cite{Giotis17,retsinas2018efficient}.
The two problems are closely connected, and both have their analogous counterparts in speech and audiovisual signal procesing \cite{petridis2018end,jha2018word}.
Yet, very often they are faced with different families of techniques.
%Methods based on some variant of Deep Neural Networks (RNN) constitute the state-of-the-art for HTR today, outperforming older ``non-deep'' methods such as Hidden Markov Model-based methods.

In this work, we present \emph{Word Spotting and Recognition Network (WSRNet)}, a unified model that can tackle both handwriting recognition and keyword spotting, using the same neural network architecture.
The proposed architecture fuses two components that are prevalent in Recurrent Neural Network-based methods, namely the Connectionist Temporal Classification (CTC)-based training and the sequence-to-sequence (Seq2Seq) paradigm.
The two components are joined in a single architecture through a suitable multi-task loss, where we show that both components are necessary for optimally efficient model training.
%\item simplify baseline arch based on ctc training.
Full word recognition is possible by either of the CTC or the Seq2Seq paths, while both the CTC and Seq2Seq network branches are used to backpropagate losses during training.
The option of keyword spotting is enabled by taking advantage of the Seq2Seq intermediate fixed sized encoding as a feature vector.
Then KWS can be performed by Example (Query-by-Example, QbE) simply by comparing feature vectors or by String (Query-by-String, QbS) by employing an extra encoder module that translates query strings to the Seq2Seq intermediate representation space, or by forced aligning the query to the decoder.
Furthermore, we show that the Seq2Seq-based representation can be refined by binarizing it with an efficient straight-through estimator-based (STE) retraining scheme \cite{bengio2013estimating}.
This binary representation, aside from being very compact and economical in terms of space, also very crucially allows for very fast KWS.
%\item Explore generated holistic representation space. -- This could become "just" a figure, later
Numerical experiments show that the proposed unified model outperforms the current state-of-the-art in KWS, creating a new baseline. 
In word recognition, the model also leads to state-of-the-art performance, very close to that of the leading method in recognition (less than $0.08\%$ difference in CER performance).

%\item explore Seq2Seq approach
%\item intermediate Seq2Seq feat vector for holistic word representation
%\item qbs with extra encoder module
%\item binarization of holistic representation with a re-training approach based on straight-through  estimator.

%\item QbS KWS by forced alignment on decoder.
%\item Significantly improve SoA results

The remainder of this paper is organized as follows.
In section \ref{sec:related} we briefly examine the related literature on handwriting recognition and keyword spotting.
In section \ref{sec:proposed} we present the proposed architecture and outline its use for word recognition.
In sections \ref{sec:kws} we discuss how to use the proposed architecture for QbE and QbS keyword spotting.
We present numerical experiments on both tasks in section \ref{sec:experiments} and conclude the paper with section \ref{sec:conclusion}.

%-------------------------------------------------------------------------
\section{Related Work}
\label{sec:related}

As is the case with most, if not all, tasks in computer vision, both handwriting recognition and keyword spotting are today dominated by neural network-based methods.
In handwriting recognition, recurrent neural networks have become the baseline, as they naturally fit with the sequential nature of handwriting, and especially after the introduction of a number of key elements to the standard recurrent network paradigm \cite{fischer2012handwriting,fischer2012lexicon}.
Such key advances include the Long Short-Term Memory model (LSTM) \cite{greff2016lstm}, that effectively dealt with the vanishing gradient problem, and the Connectionist Temporal Classification (CTC) method and corresponding output layer \cite{graves2006connectionist,graves2012connectionist}.
With CTC, a differentiable output layer that maps a sequential input into per-time unit softmax outputs, allowing simultaneous sequence alignment and recognition with a suitable decoding scheme.
% Gated recurrent units ?
Research on decoding schemes is also active \cite{collobert2019fully}, with the beam search algorithm being a popular approach, capable of exploiting an external lexicon as an implicit language model.
RNN-based approaches have thus practically overshadowed the previous state-of-the-art, which was based mostly on Hidden Markov Model (HMM)-based approaches \cite{puigcerver2017multidimensional}.
%Extension to multidimensional RNNs has also been proposed \cite{leifert2016cells}.
Multi-dimensional RNNs have been considered for HTR \cite{leifert2016cells}, however there has been criticism that the extra computational overhead may not translate to an analogous increase in efficiency~\cite{puigcerver2017multidimensional}.
Convolutional neural networks have also been employed, especially for word-level recognition \cite{krishnan2016deep,ptucha2019intelligent,poznanski2016cnn,such2018fully} where a lexicon of possible targets may typically be required.
Recent developments in CNNs for lexicon-free word recognition typically include a recurrent network component \cite{krishnan2018word,dutta2018improving,toledo2017handwriting}.
For example in \cite{toledo2017handwriting}, the convolutional PHOCNet architecture \cite{sudholt2016phocnet} is used to embed words to an attribute space, passed to a Bidirectional LSTM (BLSTM)-CTC component.
%A CNN-RNN model that can perform either lexicon-based or lexicon-free recognition.
%Included many useful quirks such as: 
%a) A spatial transformer network. 
%b) Pretraining on the (publicly available) IIT-HWS %\footnote{\url{http://cvit.iiit.ac.in/research/projects/cvit-projects/hwnet}}
%c) Extensive training-time augmentation.
%d) Test-time augmentation.
%Other recent developments: 
%Following an alternative strategy again for RNN-based recognition, the sequence-to-sequence architecture has been used for recognition \cite{sueiras2018offline}. 
%1. Rochester lab models: Ptucha et al. 2019 \cite{ptucha2019intelligent}, Such et al. \cite{such2018fully}.

%\emph{Lexicon-based methods:}
%Almaz\'{a}n et al. 2014 \cite{Almazan14PAMI}: 
%Embed words and lexicon strings to a common subspace, then assign recognized transcription as the nearest neighbor.
%Krishnan et al. 2016 \cite{krishnan2016deep} and the improved version \cite{krishnan2018word}.
%Uses a convolutional network to produce attribute-based embeddings like Almaz'{a}n et al \cite{Almazan14PAMI}. 
Regarding KWS \cite{Giotis17}, a number of recent methods have been inspired by the attribute-based model of Almaz\'{a}n et al. \cite{Almazan14PAMI}.
In this model, character-level attributes are learned as a Pyramidal-Histogram-of-Characters fixed-size vector (PHOC) and projected along with string representations to a common subspace, allowing QbE and QbS word-level KWS.
This base model has been further extended or adapted \cite{krishnan2016deep,sudholt2016phocnet} using convolutional neural networks to replace the whole or part of the pipeline. %\cite{krishnan2018hwnet}.
%All these word-based approaches for keyword spotting make use of Convolutional Neural Networks (CNNs) instead of SVMs and KCSR for the attribute and common subspace learning.
%Sudholt and Fink \cite{sudholt2016phocnet} were the first to use CNNs in order to learn the PHOC representation given the raw word image content.
%The method is able to outperform Almaz\'{a}n's method under both QbE and QbS scenarios.
In Krishnan et al. \cite{krishnan2016deep} for example, a word image representation is first learned using a CNN and subsequently used to learn a common subspace with KCSR as in \cite{Almazan14PAMI}.
Also of note is that these methods have been used for word recognition, albeit lexicon-based, where the rationale is to compare the common image attribute / string representations and return the closest match in the lexicon.
The attribute-based PHOC representation has been shown to be decodeable without use of a lexicon, with some limited success \cite{sfikas2017phoc}.
With respect to Query-by-String for line-based KWS, RNN-based models like the BLSTM+CTC-based method of Frinken et al. \cite{Frinken12} constitute here too the state-of-the-art, as in HTR methods, outperforming non-RNN based methods like HMM-based ones \cite{toselli2016hmm}.
The sequence-to-sequence architecture has led to state-of-the-art results in Natural Language Processing, involving translating an input sequence to an output sequence of a different length in general.
Use of the Seq2Seq architecture has started been used in HTR and KWS recently as well \cite{sueiras2018offline,wei2019word}.
%Also notable is the method for KWS presented in \cite{retsinas2018efficient}, where a learning-free method is proposed %will cite on intro, have to think about where to put it

%{\bf Compact representations}: Our paper "compact..."
%Sparse priors..
%Refer to straight-through estimation? 

%TODO: Refer to CNN-N-grams explicitly? It was a CVPR paper

Let us note that models that can handle both HTR and KWS with the same model mechanism have been previously proposed, with significant success \cite{Almazan14PAMI,krishnan2016deep}.
Many works, especially on line-level, deal with KWS as constrained HTR \cite{Frinken12,toselli2016hmm}.
In this paper, the two tasks also form part of the same paradigm, and the HTR task can be seen as a means to obtain more discriminative features for KWS.

%------------------------------------------------------------------------

\section{Proposed Architecture and Word Recognition}
%\section{Proposed Architecture}
\label{sec:proposed}
        
\begin{figure*}[t]
    \begin{center}
    \includegraphics[width=0.99\linewidth]{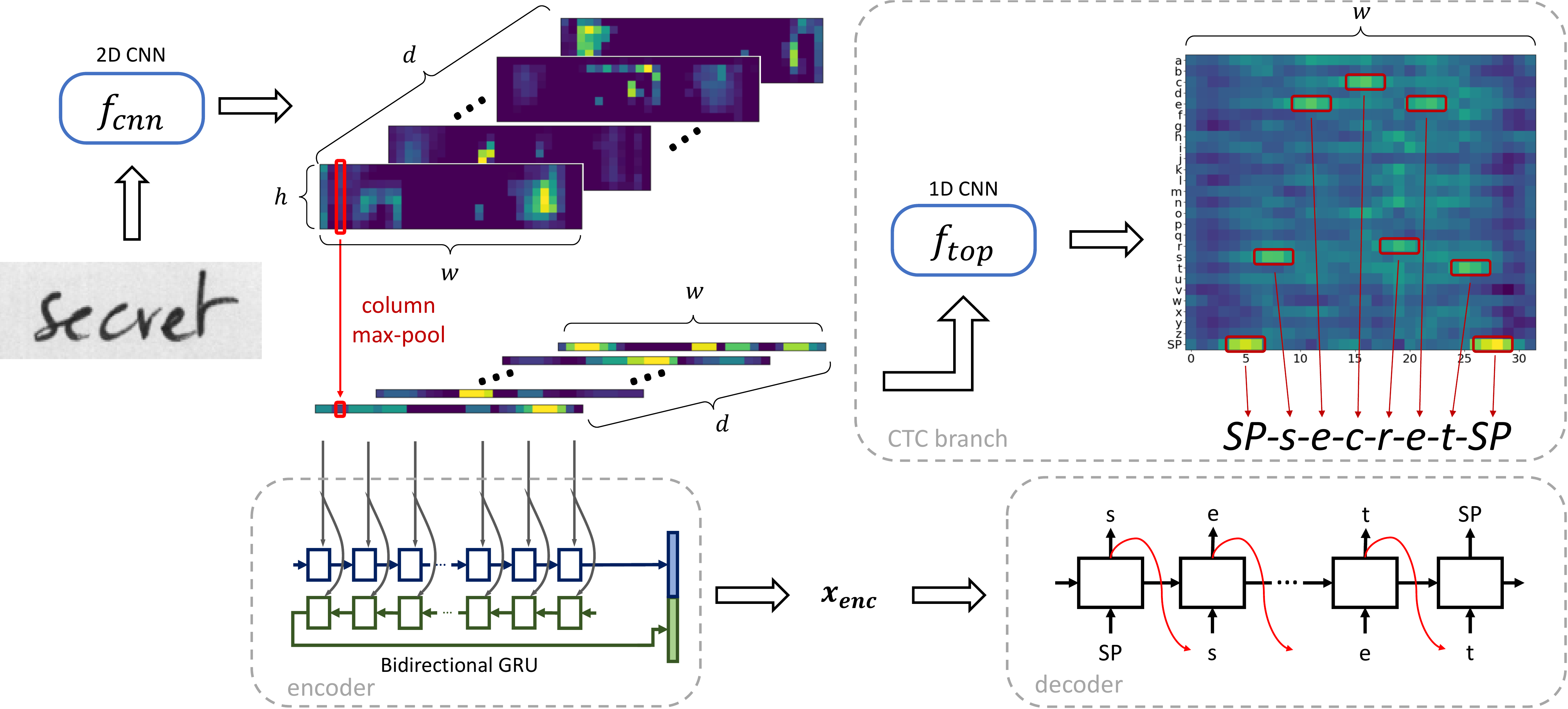}
    \end{center}
       \caption{Proposed architecture. 
       The word image is processed through a 2D convolutional backbone, leading to $w \times d$-sized feature map.
       The model then branches to two components, a CTC branch and a Seq2Seq (encoder/decoder) branch, combined through a multi-task loss.
       Omitting the ``traditional'' recurrent layer before the CTC output results in easier model training,
       while combining the two branches leads to overall increased performance for either branch 
       %(CTC,Seq2Seq) 
       and task.
       %(HTR,KWS)
       }
       %TODO: Perhaps a word or two about how to do KWS / HTR?

       %and fed to a 1D convolutional head and CTC output layer.
        %Proposed architecture Seq2Seq branch.
        %The backbone is shared with the CTC branch (Fig.~\ref{fig:overview}).
    \label{fig:overview}
    \end{figure*}

In this section, we describe in detail the proposed architecture for word recognition and spotting, first with a focus on the former application.
Our architecture comprises two basic components, namely a CTC-based branch and a Seq2Seq branch.
The CTC pipeline is inspired by typical CTC-based HTR systems applied in text-line level, as in \cite{fischer2012handwriting} %puig ??
Contrary to these systems, which consist of a CNN for feature extraction topped by a RNN (usually LSTM), we replace the RNN module with an 1D CNN.
This architectural choice aims to simplify the pipeline and assist the training procedure, as we will explain in detail in the following sections.
Over the CTC pipeline we added a Seq2Seq module \cite{sutskever2014sequence}, i.e. a Encoder/Decoder Recurrent network pair, which takes as input the output of the aforementioned CNN module, 
encodes the word information into a fixed-sized vector and consequently decodes it into a sequence of characters.
%Specifically, one important aspect of this work is to explore how the simultaneous training of the two recognition pipelines affects their performance.
The proposed model architecture can be examined in Figure~\ref{fig:overview}. 
In what follows, we describe our architecture modules and their functionality in detail.

%\subsection{CTC-based Baseline}
\subsection{Convolutional backbone}

%First, we describe the CTC subcomponents: 1) CNN Backbone and 2) 1D CNN Head.
%\subsubsection{Backbone Network}

The visual feature extraction task is performed by a CNN, dubbed as the convolutional backbone.
This part of the network produces a feature map based on a word image input, that will be subsequently processed by two ``heads'', the CTC and Seq2Seq branches.
%which consists of multiple residual blocks along with ReLU nonlinearities. % \cite{strang2019linear}. 
%CHECK:
A total of four convolutional stacks ($conv1 - conv4$), comprising $1, 2, 4$ and $4$ layers respectively, form residual blocks (all but $conv1$) topped by ReLU nonlinearities preceded by Batch Normalization and dropout layers.
Convolution window sizes on stacks $1$ to $4$ are $7 \times 7$, $3 \times 3$, $3 \times 3$, $3 \times 3$, with depths equal to $64, 64, 128, 128$ respectively.
Max-pooling is performed on $2 \times 2$ windows of stride $2$.
%To assist network convergence, we add batch normalization (BN) before nonlinearities, and use dropout at each layer~\cite{bengio2017deep}. %in order to improve generalization
%The feature map downsampling between groups of consecutive residual blocks is performed by max pooling.
%The CNN architecture is summarized on Table~\ref{table:cnnbackbone}.
%CNN backbone architecture: Convolutional stacks of residual blocks transform the input word image to a feature map, subsequently passed to the CTC Convolutional head.
To further promote simplicity we transform the output of the CNN backbone of size $h\times w \times d$ into a feature sequence of size $w \times d$ by column-wise max-pooling. 
The reasoning behind max-pooling is that we care only about the existence of features related to a character and not their spatial position. 
%Note that if column-wise concatenation was performed, instead of max-pooling, the feature sequence would be of size $w \times hd$.

\subsection{CTC component}

Typically, CTC-based approaches involve using a recurrent layer (e.g., an LSTM) as their network head.
In this work, contrary to the majority of existing CTC-based approaches, the proposed CTC branch does not comprise any recurrent layers.
Instead, a batch of three 1D convolutional layers with kernel size equal to $7$, along with BN, ReLU and dropout are used.
Note that multiple 1D convolutional layers are capable of encoding context-wise information, which is the major goal of recurrent networks.
The gain of replacing recurrent networks with 1D CNNs is two-fold:
First, LSTMs are known to exhibit convergence difficulties, while 1D convolutions with BN can converge very fast.
%ECCV GS: According to the rebuttal, 
%1D CNN does not improve the CTC component, rather helps its convergence (back-propagation on CNN layers is much "easier" than on LSTMs) and consequently the convergence of Seq2Seq branch, while being cost-efficient (no RNNs are involved).
Overall convergence is assisted by quickly generating discriminative features at the top of the CNN backbone, simplifying the Seq2Seq task.
Second, 1D convolutions can be fully parallelized and thus considerably improve training and inference time, as opposed to LSTMs.

The output of the 1D CNN is of size $w \times n_{classes}$, where $n_{classes}$ is the number of possible character classes. 
Applying the softmax function on the final output, we form a sequence of probability distribution over the possible characters which is then propagated into the CTC loss $L_{CTC}$.
Given the trained system and an input image, a sequence of character probabilities is generated and it can be eventually transformed into the recognized character sequence by a CTC decoding procedure \cite{graves2012connectionist}.
%(based on greedy or dynamic programming algorithms)~\cite{}
%\footnote{CTC includes a blank token, which should be properly treated at the decoding step.}.

\subsection{Sequence-to-Sequence Component}

The second branch of the proposed network architecture involves a sequence-to-sequence component \cite{sutskever2014sequence}.
The main components of such a system are two recurrent neural networks, the \emph{encoder}, which projects a sequence into a fixed-sized vector, and the \emph{decoder}, which is responsible for decoding the encoded fixed-sized feature vector into the target sequence.
The encoder network generates the fixed-sized vector by extracting the last hidden vector of the recurrent operation as a holistic representation.
The decoder network, given a hidden vector and the previous element of the sequence, predicts the next element. %, as described in detail in \cite{}.
Concerning the problem of handwriting recognition, the input sequence is the sequence of visual features, generated by the backbone CNN, as described at the previous section.
The output sequence is, as expected, the target sequence of characters.
Therefore, given a input character and an intermediate feature representation (hidden vector), the decoder system should predict the next character.
For our system, both the starting and ending tokens (starting and ending a word, respectively) are selected to be the same space token (SP), which naturally separates words.

%The proposed Seq2Seq branch is visualized at Figure~\ref{fig:s2s}, for which our architectural choices are as follows:
%    \item \emph{Encoder:}  bi-directional GRU  %(an alternative of LSTMs) 
%    comprising $3$ layers by $256$ hidden vector size.
%    The $2 \cdot 3 \cdot 256$ dimensional output vector is then compressed into $512$ dimensions by a linear transformation, which is the final output of the encoder and, consequently, the input of the decoder.
%    \item \emph{Decoder:} one-directional GRU of $1$ layer and 512 hidden size. 
%    The output at each step is transformed by a linear layer into $n_{classes}$ size in order to predict the next character. Since this formulation leads to a classification problem per step, a softmax is applied on each output and the widely-used cross entropy loss is selected.
%\begin{itemize}

    \emph{Encoder:}  bidirectional GRU \cite{chung2014empirical} %(an alternative of LSTMs) 
    of 3 layers with 256 hidden size. The $2 \times 3 \times 256$ dimensional output vector is then compressed into $512$ dimensions by a linear transformation, which is the final output of the encoder and, consequently, the input of the decoder.

    \emph{Decoder:} unidirectional GRU of 1 layer and 512 hidden size. 
    The output at each step is transformed by a linear layer into $n_{classes}$ size in order to predict the next character. 
    %Since this formulation leads to a classification problem per step, a softmax is applied on each output and the cross entropy loss is selected.
    Let $g_{d}$ be the GRU cell of the decoder module and $s = c_0 c_1 \cdots c_{K-1}$ the target string. Note that, according to our formulation, $c_0 = c_{K-1} = \text{SP}$, where $SP$ stands for the ``blank'' token.
    Also, let $x_{enc}$ the output of the encoder module. 
    Each decoding step can be written as:
    \begin{align}
        c_0 = \text{SP}\quad \& \quad h_0 = x_{enc} \nonumber \\
        c'_i,\, h_i = g_d(c_{i-1}, h_{i-1})
    \end{align} 
%\end{itemize}

Concerning the use of attention \cite{prabhavalkar2017analysis}, we have chosen against using it with our Seq2Seq component.
State-of-the-art Seq2Seq models do use an attention module \cite{prabhavalkar2017analysis}, which directly propagates information from the input sequence to the output sequence.
However, such an assisting module would result in decreasing the significance of the intermediate vector between the encoder and the decoder. 
Specifically, for attention-based approaches, the character encoding information is mostly propagated through the attention module while the intermediate feature vector usually assists the spatial correspondence of the attention module.
As we mentioned in the introduction, one of our main goals it to generate unique word representations. 
The intermediate feature vector is ideal for this task and thus adding an attention path would decrease the ability of generating discriminative representations.

\iffalse
\begin{algorithm}
\caption{Seq2Seq Training}\label{alg:s2s}
\begin{algorithmic}[1]
%\Procedure{S2S}{}
\State $c_0 \gets \text{SP}$
\State $h_0 \gets x_{enc}$
\State $i \gets 0$
\State $L \gets 0$
\If {teacher forcing}
    \While{$i<K$}
        \State $c^{pred}_i,\, h_i \gets g_d(c_{i-1}, h_{i-1})$
        \State $L \gets L{ce}(c^{pred}_i, c_i)$
        \State $i \gets i + 1$
    \EndWhile
\Else
    \While{$i<K+2$}
        \State $c^{pred}_i,\, h_i \gets g_d(c^{pred}_{i-1}, h_{i-1})$
        \State $L \gets L{ce}(c^{pred}_i, c_i)$
        \State $i \gets i + 1$
    \EndWhile 
\EndIf
%\EndProcedure
\end{algorithmic}
\end{algorithm}
\fi

The Seq2Seq component shares the same backbone with the CTC-based component. 
%In particular, this is the convolutional path from the input word image up to and including the max-pooling result to a feature map of size $w \times d$ (see Figures \ref{fig:overview},\ref{fig:s2s}).
The full model is then trained with a multi-task loss, defined as a weighted sum of the loss for the two branches:
\[
    L(w_{cnn}, w_{top}, w_{S2S}) =
\]
\begin{equation}
    L_{CTC}(w_{cnn}, w_{top}) + \lambda L_{S2S}(w_{cnn}, w_{S2S}),
    \label{eq:mtask}
\end{equation}
where $L_{CTC}$ and $L_{S2S}$ are the CTC and the Seq2Seq loss functions respectively and hyperparameter $\lambda$ controls the contribution of each loss.
$L_{S2S}$ loss is the average cross entropy loss across all the per-character predictions produced by the decoder.
The parameters of the backbone CNN $w_{cnn}$ are jointly trained by both losses, while the parameters of the 1D CNN $w_{top}$ and the parameters of the Seq2Seq modules $w_{S2S}$ are optimized by their corresponding losses.

In inference mode, this network branch offers an alternative to the CTC for decoding the input, by simply computing the Seq2Seq output sequence.
In practice, joint training offers an improved decoding for either of the two inference options, compared to having trained the two branches separately.

\iffalse
\begin{figure}[t]
    \begin{center}
    \includegraphics[width=0.99\linewidth]{img/s2s.pdf}
    \end{center}
       \caption{Proposed architecture Seq2Seq branch.
       The backbone is shared with the CTC branch (Fig.~\ref{fig:overview}).
       The Seq2Seq branch is combined with the rest of the network through a multi-task loss.}
    \label{fig:s2s}
    \end{figure}
\fi
    
\iffalse
\begin{figure}[t]
\begin{center}
\includegraphics[width=0.9\linewidth]{decoder.pdf}
\end{center}
   \caption{??}
\label{fig:decoder}
\end{figure}
\fi

\section{Keyword spotting using the Seq2Seq encoding} %as By-Product}
\label{sec:kws}

Apart from evaluating the Seq2Seq approach on HTR, in this section we show that the proposed architecture can also be used to tackle either QbE or QbS keyword spotting. 
Furthermore, we explore further refining the resulting holistic word representation by binarizing and thus significantly compressing it.
Regarding QbS KWS, we present two alternative schemes, one involving adding an autoencoder module to the main network, and the other involving using the technique of forced alignment on the Seq2Seq branch.

\subsection{Query-by-Example with the encoder output as a word image representation}

%So far we have build a well-performing word recognition system, but we have not yet tackled the keyword spotting problem.
%Although many approaches exist in the literature, state-of-the-art approaches share a similar trait, they extract deep features by either performing word classification (many classes - poorly generalization for out-of-vocabulary words) or attribute detection (each word is defined by a set of attributes - does not guarantee the existence of a decoding process).
The idea behind the proposed word spotting extension is to utilize the architecture to generate descriptive holistic representations for each word image and the query.
The intermediate feature vector of the Seq2Seq system, generated between the encoder and the decoder module, is ideal for this task, as it readily produces a fixed-sized descriptive word descriptor.
Query-by-Example KWS is then straightforward, as it suffices to compare descriptors with a suitable distance measure.% we only have to compute their cosine distance and thus we can perform template matching (QbE) very efficiently. 
It is also important to note that the generated descriptor can be fully translated into the target character sequence through the decoder module, as opposed to attribute-based word representations, such as Pyramidal Histogram of Characters (PHOC)~\cite{Almazan14PAMI,sudholt2016phocnet}.

\subsection{Query-by-String with an Autoencoder Module}
\label{subsec:autoencoder}

The existing system can straightforwardly perform QbE spotting by comparing the generated intermediate feature vectors.
Nevertheless, the QbS variation cannot be executed with the current formulation.
To this end, we add an extra encoder module which encodes the target sequence (i.e. in practice, the string query) into a feature vector
which can be decoded with the existing decoder into the same sequence, effectively forming an autoencoder path.
Therefore, with this autoencoder we construct a one-to-one function between character sequences and intermediate fixed-sized feature vectors.
%Therefore each word has a unique representation.

The problem that arises from the addition of the extra character encoder module is the simultaneous training along with the visual encoder of the Seq2Seq component.
If we constrain the output of the character encoder to be similar to the visual encoder output, e.g. using MSE loss, 
there is no guarantee that both encoding can be decoded successfully, since even a small Euclidean divergence between the encoding may result into two different decodings.
We overcome this problem by randomly choosing to decode one of the two information flows at each iteration, while constraining them to be close to each other.
More formally, let $f_{cnn}$ be the backbone CNN, $f_{enc}$/$f_{dec}$ the encoder/decoder of Seq2Seq and $f_{cenc}$ the character encoder. 
Given an image $I$ and its corresponding groundtruth string $s = c_0 c_1 \dots c_K$, the two encodings can be written as follows:
\begin{align}
    x_{enc} &= f_{enc}(f_{cnn}(I; w_{cnn}); w_{enc})\\
    x_{cenc} &= f_{cenc}(s; w_{cenc})
\end{align}
The loss for the Seq2Seq branch is then extended as:
\begin{equation}
    \label{eq:s2sloss_new}
L_{S2S}(s_{pred}, s; w_{cnn}, w_{S2S}) + d_{cos}(x_{enc}, x_{cenc})
%\frac{x_{enc} \cdot x_{cenc}}{\| x_{enc} \|\, \|x_{cenc}\|},
\end{equation}
where $d_{cos}(\cdot,\cdot)$ stands for the cosine distance, and the predicted string $s_{pred}$ is the decoding of either $x_{enc}$ or $x_{cenc}$, randomly selected, as follows:  
\begin{gather}
    b  \sim Bernoulli(0.5)\\
    s_{pred}  = f_{dec}(b  x_{enc} + (1-b) x_{cenc}; w_{dec})
\end{gather}

\iffalse
The aforementioned training scheme is summarized at Figure~\ref{fig:autoencoder}.
\begin{figure}[t]
\begin{center}
\includegraphics[width=0.75\linewidth]{autoencoder.pdf}
\end{center}
   \caption{Autoencoder module that extends on the Seq2Seq component. 
   By learning a mapping from a character sequence to the Seq2Seq intermediate representation, Query-by-String KWS is effectively enabled.
   }
\label{fig:autoencoder}
\end{figure}
\fi

During model training, this extended loss (Eq.~\ref{eq:s2sloss_new}) replaces the corresponding term in the full model multi-task loss (Eq.~\ref{eq:mtask}).
% ECCV: Esvisa to parakatw, symfwna me to rebuttal rev1.2
%Aside from enabling QbS, the addition of the autoencoder module is beneficial implicitly also for the QbE scenario, as it effectively results in constraining 
representations of different word images depicting the same word around a unique representation.

An additional interesting consequence of adding the autoencoder module is that a word corpus (a collection of word transcripts) can be used to aid training.
%\emph{external} lexicon (\emph{external} in the sense that it contains word transcripts that are not necessarily part of the training set, nor its words are paired with ground-truth images) 
So far, we took into account only the words exist in the training set, even though Seq2Seq system is capable of learning an implicit language model of valid consecutive characters. %(this would be further highlighted into the experimental section).
The intuition behind this variation is to assist the underlying implicit language model of the Seq2Seq system, by feeding it with valid words that may not exist in the training set.
Implementation-wise, at each optimization iteration and after updating weights by backpropagating the multi-task loss w.r.t. the standard training set, we fine-tune %(using a small learning rate) 
the autoencoder path with words drawn from the word corpus.

\subsection{Query-by-String with Forced Alignment} %: An Alternative to QbS KWS}
\label{subsec:fd}

%\textcolor{red}{force decoding??}
%\textcolor{red}{Sfikas: check notation on decoder, autoencoder \& here}

Even though holistic representations of fixed size greatly simplify the upcoming matching step, we also consider the case of using the decoder model of the proposed Seq2Seq system as a scoring function for the representation $x_{enc}$ constrained to produce a target word string $s$.
This constrained decoding, usually referred to as forced alignment, is a popular QbS alternative based on character lattices~\cite{toselli2016hmm}, where a specific sequence of character is scored according to a pre-computed character graph (consisted of the probability of the character at each node, as well as the transition probability from one node to another).
The alignment term refers to the case of several possible alignments of the sequence of features to the desired sequence of characters.
However, following the decoder definition, the query-constrained scoring can be efficiently performed as the following equations suggest: 
\begin{gather}
    c_0 = \text{SP}\quad \& \quad h_0 = x_{enc} \nonumber \\
    c'_i,\, h_i = g_d(c_{i-1}, h_{i-1}), \quad i=1,\dots,K-1 \nonumber \\
    \text{score} = \sum_{j=1}^{K-1} L_{CE}(c'_{j}, c_{j})
    \label{eq:force_align}
\end{gather} 
Specifically, we assume that the input of the decoder is the requested query $s=c_0 c_1 \dots c_{K-1}$ and thus it is straightforward to predict the next character, given the previous one and the hidden vector computed this far.
Consequently, the score is the average cross entropy loss of the predictions and thus if the score is low, the given query was in line with the word representation $x_{enc}$.

Implementation-wise, performing a forced alignment as described above is time-consuming and cannot be parallelized in order to fully utilize accelerators, e.g. GPUs.
We can however overcome the computational cost by organizing multiple queries into a character trie:
given a single intermediate feature, we decode over the character trie in a breadth-first traversal manner, making use of multiple nodes for a parallelized, fast implementation on the GPU.

\iffalse
Given a target word $W=c_0 c_1 \dots c_{n_w}$, an intermediate feature vector $x_{enc}$
\begin{align}
    h_{t} = f_{gru}(h_{t-1}, c_{t}), \quad h_0 = x_{enc}, \quad t=1,...,n_w\\
    p_{t}= f_{h2c}(h_{t})\\
    \text{score} = \sum_{i=1}^{n_w} L_{CE}(p_{t}, c_{t})
    \label{eq:force_align}
\end{align}
\fi

\subsection{Binary Word Representation}
\label{subsec:binaryrep}

One noteworthy variation is the binarization of the intermediate feature vector, which can greatly reduce the storage requirements for storing large collections of documents.
For example, if we have a collection comprised of $1,000$ words (assuming perfect word segmentation), the proposed representation has a dimensionality of $512$ with floating-point values, which amounts to $512 \times 4$ bytes and overall almost $2.5$ GB for the whole collection. 
On the other hand, by assuming a binary $512$-dimensional representation, the overall storage requirements are only $64MB$, namely $32\times$ less storage.
The binarization of the word representation can be simply performed by a \emph{sign} operation and we expect to generate binary descriptors that do not affect significantly the performance of KWS task, since the already used cosine distance would also work well under the signed vector, as the experimental results validate.
Note that all the signed vectors have the same magnitude and therefore, since we care about comparing an image representation with a reference one, the cosine similarity can be efficiently computed by \emph{XNOR} and \emph{bincount} operations.
Nevertheless, such a binarization approach may significantly affect the decoding step and thus the decoding property, one of the crucial contributions of this work, is lost.
To address this problem we proposed a training scheme for binarized vectors based on the Straight-Through Estimator~\cite{bengio2013estimating}.

%\subsubsection{Straight-Through Estimator}
%\label{sec:ste}

The Straight-Through Estimator was initially motivated by the inability to train networks which include threshold operations, such as the Heaviside or the sign function.
Such discontinuous functions are not differentiable and thus unsuitable for backpropagation.
To overcome this problematic behavior of threshold operations, it was proposed \cite{bengio2013estimating} to assume that these hard thresholding functions, which are used for the forward pass, let the error to be propagated without any change through at the backward pass (i.e. straight through estimation), as if the function was $f(x) = x$ and consequently had an identity gradient. 

Although STE naturally is essential for training binary CNNs~\cite{rastegari2016xnor}, it is rather crude to be applied straightforwardly on the generated representations
(we have tried such an approach, and the system could not converge).
One main difference is that STE was applied on the network weights that are steadily updated, while we apply STE to a feature vector with considerable variations for each image/word at the same iteration.
This problem was resolved by simulating a \emph{tanh} activation based on the following observation:  
%\vspace{-0.5cm}
\begin{align}
    x_{bin} = \text{sign}(x_{enc}) = \lim_{a\to\infty}\text{tanh}(a x_{enc}) 
    \label{eq:tanh_approx}
\end{align}
Therefore, we distinguish two cases:
%\begin{itemize}
%\item
In the forward pass, we simply use the binarized vector $\text{sign}(x_{enc})$ as input to the decoder;
%\item
in the backward pass, we treat the signed vector as if a \emph{tanh} operation was applied, namely $tanh(x_{enc})$, 
and compute the gradients and backpropagation error accordingly. 
%\end{itemize}
Following the STE reasoning, we do not care about defining the $a$ value of Eq.~\ref{eq:tanh_approx}, as long as we distinguish one hard operation at forward pass and a corresponding soft differentiable one for the backward.

\section{Experiments}
\label{sec:experiments}

In this section, we present experimental results across all HTR and KWS variations discussed in the previous sections.
%highlighting the contributions of this work and their impact on the system performance.
We have run numerical trials on the IAM dataset \cite{Giotis17}, consisting of a total of $13,353$ handwritten lines or $115,320$ words, written by $657$ writers, and for which both line and word level segmentations are available.
As IAM is a large and multi-writer dataset, it is typically used as the standard benchmark of comparison for either handwriting recognition or keyword spotting methods alike \cite{Giotis17,dutta2018improving,krishnan2018hwnet}.
To ease numerical evaluation comparison, a standard partition into training, test and validation subsets is also available, which is the partition used in this work.
As evaluation metrics, we use the standard metrics that are used in the related literature.
These are namely the Character Error Rate (CER) and Word Error Rate (WER) for recognition, and mean average precision (MAP) for keyword spotting \cite{Giotis17}.
CER is computed using the Levenshtein distance between estimate transcription and ground truth, while WER considers an error if the two transcriptions do not match exactly.
MAP considers retrieval precision over multiple recall thresholds, taken as a mean over the totality of the queries considered.

%One typical preparatory step in a deep learning system is the data preprocessing (e.g. normalization) and data augmentation (introducing variations/noise to synthetically enlarge the existing set of images). 
In order to train our network, we have used the ADAM optimizer \cite{strang2019linear} (initial learning rate: 0.01 / number of epochs: 80/ cosine annealing scheduler). %GRETSI: Any other details here? eg learning rate, etc.
We have augmented the training set by performing random global affine transformations on available word images \cite{sudholt2016phocnet}.
Regarding image preprocessing, we apply zero padding of all inputs to a fixed size without affecting the aspect ratio ($64\times256$). %GRETSI: which size?
This simple preprocessing enables batch processing, which leads to faster and more efficient training.
Concerning the multi-task loss weight $\lambda$ in Eq.~\ref{eq:mtask}, we found that applying a larger weight to the Seq2Seq branch ($\lambda=10$) was beneficial, as the CTC branch can be trained more easily, due to being completely non-recurrent.
Regarding training of the Seq2Seq in particular, we employ the \emph{teacher forcing scheme} \cite{bengio2015scheduled}.
Specifically, we randomly select the decoder input at step $i$ to be either the predicted character of the previous step $i-1$ or the real character $c_{i-1}$.
In this manner we avoid frequent error propagation from a miss-predicted character during step-by-step decoding. 
%The decoding procedure can be formulated as follows. %Given an the input character $c$ and a hidden vector $h$, the decoder's recurrent function $f_d$ works as follows:

%\subsection{Word Recognition Trials}

First, we explore the word recognition task with the full set of existing characters, i.e. lowercase/uppercase letters, digits and punctuation, following the line-level recognition setting.
The results are summarized at Table~\ref{table:htr}, considering both the CTC and the Seq2Seq approach. 
We distinguish the training scheme to \emph{alone}, where each pipeline is trained separately, and \emph{joint}, where both recognition flows are trained together as a multi-task problem (see Eq.~\ref{eq:mtask}). 
All the results are based on a greedy decoding procedure.
Experimental results suggest that the Seq2Seq approach has convergence difficulties when trained alone.
%\footnote{Different optimizer hyperparameters and training over a larger number of epochs may help the alone Seq2Seq solution, but it is out of the scope of this work}.
Nevertheless, when used alongside the CTC, Seq2Seq has a notable increase in performance for the exact same number of epochs.
This is in line with our initial suggestion of a fast converging CTC pipeline, which helps the generation of meaningful visual features at the CNN output in only a few epochs. % and thus providing a good "initialization" for the Seq2Seq input.
Moreover, we should highlight that Seq2Seq variation gives the best WER, while assists the CTC flow to have a minor improvement at WER too. 
\begin{table}[h]
\begin{center}
\begin{tabular}{|c|c|cc|}
 \hline
setting & branch & CER & WER \\  %ECCV: was setting & method
\hline
\hline
line & CTC \cite{puigcerver2017multidimensional} & 6.2 & 20.2\\
\hline
\multirow{2}{*}{word (alone)} 
 & CTC & 6.6 & 17.4 \\
 & Seq2Seq & 10.7 & 21.6 \\
 \hline
\multirow{2}{*}{word (joint)}  
 & CTC & 6.7 & 17.2 \\
 & Seq2Seq & 6.9 & 16.5\\
  \hline
\end{tabular}
\end{center}
\caption{Recognition results using the proposed method (setting ``word''), comparing use of the two branches (CTC or Seq2Seq).
Results are also compared w.r.t training each branch separately (``alone'') versus training w.r.t. the proposed loss.
CTC-based results on line-level recognition are also reported for comparison.}
\label{table:htr}
\end{table}
At Table~\ref{table:htr} we also report results for a state-of-the-art line-level method (CNN+LSTM) with otherwise the same trial setting (training/test set and character set) \cite{puigcerver2017multidimensional}.
Line-level segmented images are benefitted with more context compared to word-level segmentations, which can be very helpful when employing an LSTM network. 
However, word-level recognition relies on a perfect word segmentation, which is ideal, and thus avoids errors by misaligned words. 
Even though trade-off is reflected by the results, we can conclude that the proposed architectures have close performance to a typical state-of-the-art line-level recognition system. 

To further explore the capabilities of the proposed recognition system, we report the results of a more complex decoding scheme, the beam search algorithm, as shown at Table~\ref{table:beam}. 
Beam search over the CTC output is fairly common, especially for incorporating external Language Models (LMs). 
Note that no LM is used for the results.
The beam search algorithm can be easily modified for the Seq2Seq variation and specifically the decoder module. 
Instead of having a unique decoding path, where we predict the next character, we have a set of beams, each one with a different ongoing hidden vector and a history of previous characters. 
A new beam is created by adding the top-$k$ predicted characters at each step.
The beams are sorted according to the summed probability of the added characters and pruned to a predefined beam width, in order to avoid exploding complexity.
The results (we use jointly trained outputs and beam width = $5$) indicate a similar consistent error decrease for recognition with either of the model branches (CTC, Seq2Seq) when using the beam search algorithm.
In Table~\ref{table:decodings}, a number of decoding errors is shown, and the result using the two model branches is compared qualitatively.
We can deduce from these errors, which are typical in a sense of the type and difference in error between the two branches, that the two branches almost consistently lead to somewhat different decodings, even though in total their error quantitative statistics are similar.
Furthermore, the Seq2Seq errors show that this branch has learned a language model to an extent, which coincides with our expectation due to training of Seq2Seq with an implicit language model (cf. sec.~\ref{subsec:autoencoder}).

\begin{table}[h]
\begin{center}
\begin{tabular}{|c|c|cc|}
 \hline
decoding & branch & CER & WER \\ %ECCV: was 
\hline
\hline
\multirow{2}{*}{greedy}  
 & CTC & 6.7 & 17.2 \\
 & Seq2Seq & 6.9 & 16.5 \\
 \hline
\multirow{2}{*}{beam search}  
 & CTC & 6.6 & 16.9 \\
 & Seq2Seq & 6.8 & 16.2 \\
  \hline
\end{tabular}
\end{center}
\caption{
    Comparison of decoding schemes for word recognition.
    Beam search is compared to a greedy decoding scheme.
    }
\label{table:beam}
\end{table}

\begin{table}[h]
    \begin{center}
    \begin{tabular}{|c||c|c|c|c|c|}
     \hline
    ground truth & $security$ & $just$ & $laughter$ & $no$ & $roads$ \\
    \hline
    \hline
    CTC &     $seemrite$ & $jurl$ & $laughtes$ & $n0$ & $souds$ \\
    Seq2Seq & $securitte$ & $jurst$ & $laughters$ & $no$ & $souch$ \\
    \hline
    \end{tabular}
    \end{center}
    \caption{
        Error decoding examples using the proposed model.
        Results using either model branch are compared (CTC/Seq2Seq).
        }
    \label{table:decodings}
\end{table}
    
\iffalse
%GRETSI: Tha mporousan isws afta na mpoun se ena pinakaki... 
%(?) den kserw an aksizei ton kopo/an prosferei kati sto paper (to sxetiko sxolio gia future work to piga sto conclusion)
\textcolor{red}{Examples:
orig::  security 
gdec::  seemrite 
adec:: securitte /
orig::  just 
gdec::  jurl 
adec:: jurst /   
orig::  laughter 
gdec::  laughtes 
adec:: laughters / 
orig::  no 
gdec::  n0 
adec:: no /
orig::  roads 
gdec::  souds 
adec:: souch  
}
\fi
%Further error analysis shows that almost consistently different decoding mistakes are observed for the two network flows.
%Based on the aforementioned observations, a beam search approach that uses both flows may increase the overall performance, but it is out of the scope of this work and is left for future research.

%\subsection{Keyword spotting trials}

%ECCV: TODO: "LM helps to better generalize to OOV words... (apo rebuttal)"
Having established the functionality of the recognition system, we focus on exploring the proposed variations for the keyword spotting task.
Spotting approaches usually use a different character set, consisted only of lowercase letters and digits, and thus we follow the same setting, both for word spotting as well as the reported recognition results from now on. 
Table~\ref{table:kws} contains the experimental results for both QbE and QbS, along with the recognition metrics for the decoder. 
The system under consideration comprises both the CTC and Seq2Seq branches, as well as the autoencoder module.
Recognition is performed using the Seq2Seq branch. %, since the proposed intermediate representation 
This system is dubbed as \emph{WSRNet}, while the variation of employing an external word corpus in order to learn an implicit Language Model is referred to as \emph{WSRNet+LM}.
We learn the implicit LM using a concatenation of the LOB \cite{johansson1986tagged} and BROWN \cite{francis1964brown} corpora, from which we sample words according to their occurrence frequency.
We also evaluate two distinct training schemes: we compare training from scratch the whole network versus 
using the previously trained modules for the recognition task as initialization then fine-tuning the autoencoder. %GRETSI: Check, is that it? 
%(fine-tuning approach, even though the character encoder module is new). 
The experimental results indicate that the LM variant improves the decoding process by a considerable margin, while having a positive impact on KWS as well (especially for the case of training from scratch).
Moreover, the initialization of the majority of the system weights with the previously trained recognition task leads to a noteworthy boost.

\begin{table}[h]
\begin{center}
\begin{tabular}{|c|cc|cc|}
 \hline
method & QbE & QbS & CER & WER \\
\hline
\hline
\multicolumn{5}{|c|}{from scratch}\\
\hline
WSRNet  & 90.20 & 92.07 & 6.5 & 17.2 \\
WSRNet+LM & 90.87 & 93.16 & 5.9 & 15.8\\
\hline
\multicolumn{5}{|c|}{fine-tuned}\\
\hline
WSRNet & 92.09 & 95.23 & 5.1 & 14.1\\
WSRNet+LM & 92.05 & 95.67 & 5.0 & 13.9\\
\hline
%other & & & &
\end{tabular}
\end{center}
\caption{
    Comparison of recognition and KWS results for incorporating a language model.
    Recognition on the Seq2Seq is considered, and QbS is performed using the autoencoder module.
}
\label{table:kws}
\end{table}

Next, we present trials concerning the binarization strategy of the proposed word representation over the fine-tuned \emph{WSRNet+LM} variation. 
Table~\ref{table:binarization} summarizes the effect of the binarization.
Perhaps unsurprisingly, applying the \emph{sign} operator on the intermediate feature vector and without retraining the network, with impressive spotting results (only $\sim 1 \%$ drop in both metrics) and very poor recognition performance. 
However, by performing the retraining strategy with the STE, as proposed in subsection~\ref{subsec:binaryrep}, we get a well-performing system over all tasks despite the information loss due to the binarization step, while at the same time obtaining a very compact descriptor ($512$-bit long).

\begin{table}[h]
\begin{center}
\begin{tabular}{|c|cc|cc|}
 \hline
method & QbE & QbS & CER & WER \\
\hline
\hline
w/o training & 90.99 & 94.11 & 92.5 & 86.9\\
w/ training & 91.31 & 93.69 & 5.7 & 15.4 \\
\hline
%other & & 
\end{tabular}
\end{center}
\caption{KWS (MAP) and recognition results using the proposed binary word representation.}
\label{table:binarization}
\end{table}

Moreover, we evaluated the efficiency of the proposed word representation with an alternative method.
We have computed statistics over the edit distance between words in our corpus and the corresponding statistics when using cosine distance on either the proposed \emph{binary representation} or the \emph{PHOC} representations ($level 5$ unigrams).
%The distance of the statistics of the proposed and PHOC descriptors against the edit distance.
We used the Kullback-Leibler divergence (KL) to express the statistics' correlation between the reference edit distance and the considered representations, and found the two divergences to be $0.0224$ (proposed), $0.0417$ (PHOC).
Hence, the proposed binary representation is closer to the desirable statistics, as expressed by edit distances between corpus words, compared to the state-of-the-art PHOC descriptor.

We have also compared the two proposed QbS/KWS strategies. 
The system under consideration is the fine-tuned \emph{WSRNet+LM} and we distinguish the use of the binarization strategy, as shown at Table~\ref{table:fd}.
We compare QbS using the Autoencoder module (section~\ref{subsec:autoencoder}) versus the Forced Alignment approach (section~\ref{subsec:fd}).
%Along with the forced alignment approach, we also report the performance of the basic QbS approach for the corresponding system, i.e. using the cosine distance between the word representations (\emph{cos}).
Notably, the Forced Alignment approach not only increases spotting performance, which was expected, but also leads to a slightly better result when used with the compact binarized vector.
Of course, the improvement is achieved at the cost of computational effort, since Autoencoder-based QbS relies on comparing fixed-length representations with a simple distance metric, and is thus very efficient.
\begin{table}[h]
\begin{center}
\begin{tabular}{|c|c|c|}
 \hline
method & binarization &  MAP\\
\hline
\hline
%\hline
\multirow{2}{*}{Autoencoder module} & No & 95.67\\
 & Yes & 93.69\\
\hline
\multirow{2}{*}{Forced Alignment} & No & 96.27\\
 & Yes & 96.33\\
 \hline
\end{tabular}
\end{center}
\caption{
    Comparison of different approaches for QbS KWS.
    The two methods for QbS integration are compared, while using or not the descriptor binarization scheme.
}
\label{table:fd}
\end{table}

\iffalse

%GRETSI: Ti ginetai me afta ???

\subsection{Encoded Feature Space Properties}
\begin{itemize}
    \item t-sne visualization. full system. 
    On queries encoding: original vs binarized
    images encoding vs queries encoding: (original only)
    \item 
    KL divergence between queries encoding and edit distance on queries/ 
    compare to that of PHOC!!
\end{itemize}
\fi

%\subsection{Comparison to State-of-the-Art}

 \begin{table}[!htb]
   \center{
   \begin{tabular}{|c|c|c|}
    \hline
     Method & CER & WER \\ \hline
%    \multicolumn{3}{|c|}{Lexicon-free methods} \\
   \hline
    Sueiras et al. \cite{sueiras2018offline} & 8.8 & 23.8 \\ 
    Wigington et al. \cite{wigington2017data} & 6.07 & 19.07 \\ 
    Krishnan et al. \cite{krishnan2018word} & 6.34 & 16.19 \\ 
    Dutta et al. \cite{dutta2018improving} & 4.88 & 12.61 \\ \hline
    \multicolumn{3}{|c|}{Proposed Models}\\ \hline
    WSRNet+LM & 5.0 & 13.9 \\
    WSRNet+LM (binarized) & 5.7 & 15.4 \\
    WSRNet+LM+beam & 4.8 & 13.6 \\
    WSRNet+LM+beam (binarized) & 5.1 & 15.0 \\
    \hline

      %Dutta, Krishnan do not use punctuation neither capitals
     \end{tabular}
     \vspace{0.1cm}
     \caption {\label{table:wer_cer} \small{
         Comparison of the state of the art for word recognition versus variations of the proposed method. 
         The greedy decoding variation versus beam search-based variation results are reported (+beam).
         All figures are CER and WER percentages computed for the IAM dataset.
         }
         }
     %GRETSI: Any news here?
   }
 \end{table}

 \begin{table}[!t]
   \center{
   \begin{tabular}{|c|c|c|}
    \hline
     Method & QbE & QbS \\ \hline \hline
%    \multicolumn{3}{|c|}{Lexicon-free methods} \\
    Attributes+KCSR~\cite{Almazan14PAMI} & 55.73 & 73.72\\
    PHOCNet~\cite{sudholt2016phocnet} & 72.51 & 82.97 \\
    HWNet~\cite{krishnan2016matching} & 80.61 & - \\
    Triplet-CNN~\cite{wilkinson2016semantic} & 81.58 & 89.49 \\
    PHOCNet-TPP~\cite{sudholt2017evaluating} & 82.74 & 93.42 \\
    DeepEmbed~\cite{krishnan2016deep} & 84.25 & 91.58 \\
    Zoning Ensemble PHOCNet~\cite{retsinas2018exploring} & 87.48  & - \\
    End2End Embed~\cite{krishnan2018word} & 89.07 & 91.26\\
    DeepEmbed~\cite{krishnan2018word}  & 90.38 & 94.04\\
    Synth+DeepEmbed~\cite{krishnan2018word} & - & 95.09 \\
    HWNetV2\cite{krishnan2018hwnet} & 90.65 & - \\
    \hline
    \multicolumn{3}{|c|}{Proposed Models}\\
     \hline
    WSRNet+LM & 92.05 & 95.67 \\
    WSRNet+LM (binarized) & 91.31 & 93.69 \\
    WSRNet+LM+FA & - & 96.27 \\
    WSRNet+LM+FA (binarized) & - & 96.33 \\
    \hline
     \end{tabular}
     \vspace{0.1cm}
     \caption {\label{table:wer_cer} \small{
        Comparison of the state of the art for keyword spotting versus variations of the proposed method.
        Autoencoder based vs Forced Alignment (+FA) QbS results are reported.
        Figures are MAP percentages computed on the IAM dataset.
        }}
   }
 \end{table}

Finally, we present a comparison of our method versus state-of-the-art methods for KWS and Word Recognition in tables \ref{table:kws} and \ref{table:wer_cer} respectively.
In KWS, we have the best results while having a very compact descriptor at the same time.
In recognition, note that while our results are marginally very close to the leading method of Dutta et al. \cite{dutta2018improving}, % ($0.12\%$ difference in CER from Dutta et al.), 
they use methods such as extensive data augmentation with local deformations and pretraining with synthetic datasets, none of which is part of the proposed pipeline.
Also note that our network, including all its subcomponents, is considerably smaller compared to vast VGG-based models such as PHOCNet~\cite{sudholt2016phocnet,sudholt2017evaluating}.

%Local deformations as an augmentation scheme as well as considerably enlarging the dataset by synthetic rendered images may benefit the performance of our system, as done in \cite{dutta2018improving}, but is out of the scope of this work.
%Refer that: \cite{dutta2018improving} also uses some other quirks like: augmentation using synthetic dataset and local deformations.

%TODO: Add CTC vs Seq2Seq decodings, Seq is better because LM
%TODO: Add ref to parameter number

\vspace{0.0cm}
\section{Conclusion}
\label{sec:conclusion}

We have proposed a novel neural network-based model that can handle either word recognition or keyword spotting.
A number of extensions and variants of the base architecture have been also proposed and discussed, 
including a retraining scheme that can produce binarized, compact descriptors for fast KWS as well as two alternative ways to handle QbS KWS with our model.
In almost all of the variations of the tasks considered, the proposed model was shown to outperform all competition with numerical trials on the IAM dataset.
For future work, we consider integrating novel layers and architectures and researching on how to build more compact networks without sacrificing efficiency \cite{zhao2019variational}.
%Also, since our error analysis in decoding results has shown that almost consistently different decoding mistakes are observed for the two network flows (CTC, Seq2Seq),
%experimentation with a beam search approach that would use both flows could increase overall recognition performance. 
%but it is out of the scope of this work and is left for future research.

%\clearpage
\bibliographystyle{unsrt}  
\bibliography{egbib}

\end{document}